\documentclass[10pt,twocolumn,letterpaper]{article}

\usepackage{cvpr}              

\usepackage[accsupp]{axessibility}

\usepackage[dvipsnames]{xcolor}

\usepackage{kotex}

\definecolor{cvprblue}{rgb}{0.21,0.49,0.74}
\usepackage[pagebackref,breaklinks,colorlinks,citecolor=cvprblue]{hyperref}

\usepackage{multirow}
\usepackage{multicol}
\usepackage{graphicx}
\usepackage{microtype}

\def\paperID{2993} 
\def\confName{CVPR}
\def\confYear{2024}

\title{Depth Prompting for Sensor-Agnostic Depth Estimation}

\author{Jin-Hwi Park\textsuperscript{\rm 1}, Chanhwi Jeong\textsuperscript{\rm 1}, Junoh Lee\textsuperscript{\rm 2} and Hae-Gon Jeon\textsuperscript{\rm 1,2}\thanks{Corresponding author}\\
\textsuperscript{\rm 1}AI Graduate School, \textsuperscript{\rm 2}School of Electrical Engineering and Computer Science, GIST, South Korea\\
{\tt\small \{jinhwipark,chanhwij,juno\}@gm.gist.ac.kr, haegonj@gist.ac.kr}}

\begin{document}
\maketitle
\begin{abstract}
Dense depth maps have been used as a key element of visual perception tasks. There have been tremendous efforts to enhance the depth quality, ranging from optimization-based to learning-based methods. Despite the remarkable progress for a long time, their applicability in the real world is limited due to systematic measurement biases such as density, sensing pattern, and scan range. It is well-known that the biases make it difficult for these methods to achieve their generalization. We observe that learning a joint representation for input modalities (e.g., images and depth), which most recent methods adopt, is sensitive to the biases. In this work, we disentangle those modalities to mitigate the biases with prompt engineering. For this, we design a novel depth prompt module to allow the desirable feature representation according to new depth distributions from either sensor types or scene configurations. Our depth prompt can be embedded into foundation models for monocular depth estimation. Through this embedding process, our method helps the pretrained model to be free from restraint of depth scan range and to provide absolute scale depth maps. We demonstrate the effectiveness of our method through extensive evaluations. Source code is publicly available at \url{https://github.com/JinhwiPark/DepthPrompting}.

\end{abstract}

\begin{figure}[t]
\centering
 \includegraphics[width=1\columnwidth]{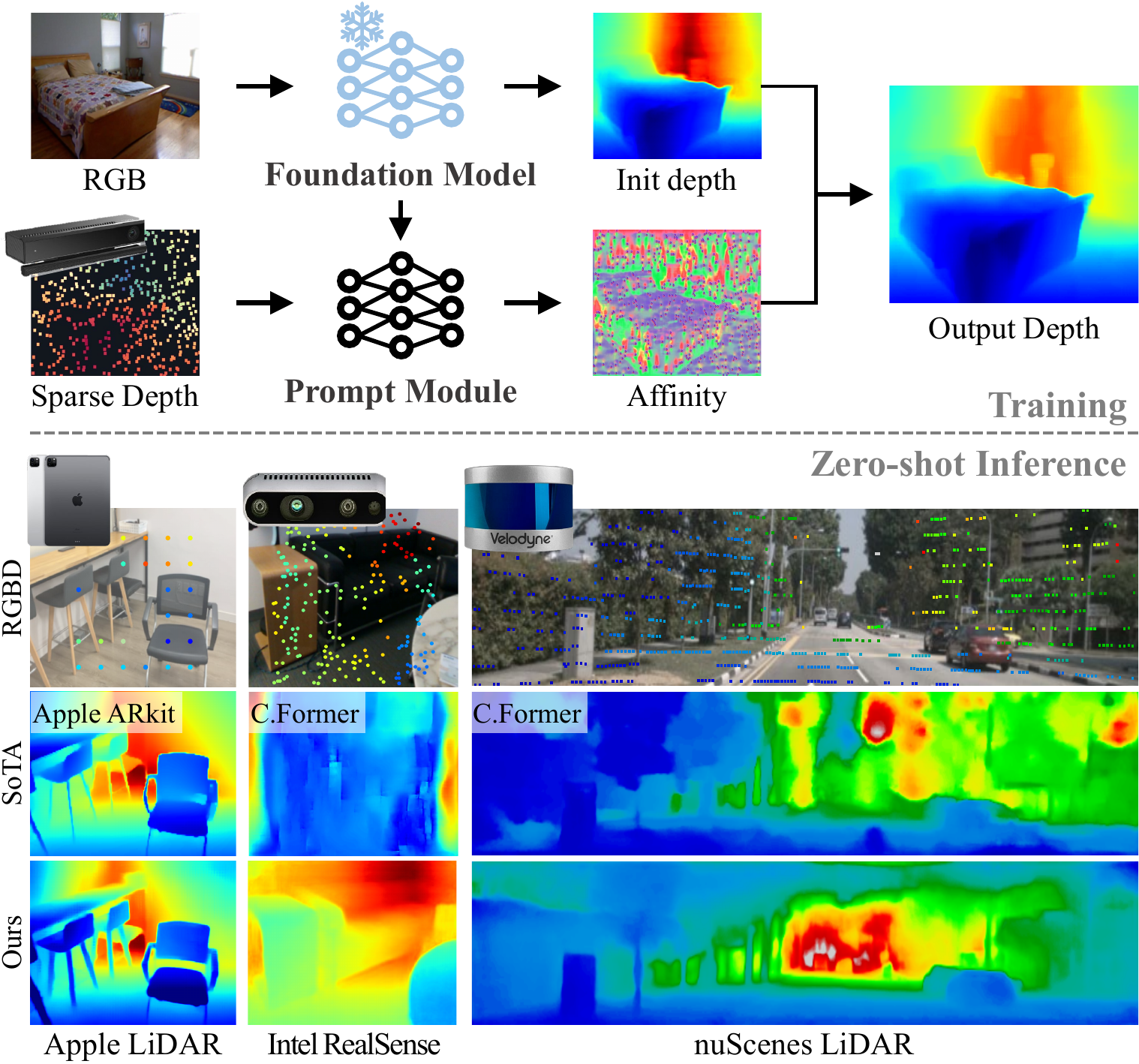}
 \vspace{-7.5mm}
 \caption{An overview of our depth prompting for sensor-agnostic depth estimation. Leveraging a foundation model for monocular depth estimation, our framework produces a high-fidelity depth map in metric scale and provides impressive zero/few-shot generality. C.Former indicates CompletionFormer~\cite{zhang2023completionformer}. More details and examples are reported in \cref{subsec: Zero-shot Inference on Commercial Sensors} and supplementary materials.}
 \label{fig:teaser}
\end{figure}

\section{Introduction}
\label{sec:intro}


Scene depths have been used as one of the key elements for various visual perception tasks such as 3D object detection~\cite{Wu2022Sparse}, action recognition~\cite{Xia2012View}, and augmented reality~\cite{Izadi2011KinectFusion,Zhou2019Learning}, etc. For accurate depth acquisition, there have been various attempts in the computer vision field. Since the advances in deep learning, its powerful representational capacity has been applied to explain scene configurations, which is feasible even with only single images. Unfortunately, single image depth estimation cannot produce metric scale 3D depths when camera parameters change and out-of-distributions on unseen datasets happen~\cite{Guo2018Learning}.

Toward depth information easy to acquire in metric scale, active sensing methods such as LiDAR (Light Detection and Ranging)~\cite{schwarz2010mapping}, ToF (Time of Flight)~\cite{lange2000demodulation}, and structured light~\cite{zhang2018high} have gained interest as a practical solution. Although the active sensing methods enable real-time scene depth acquisitions in a single shot, they only provide sparse measurements. For dense predictions, spatial propagation, modeling an affinity among input image pixels, is necessary~\cite{LearningAffinitySPN,cspn,dyspn,nlspn,park2023learning,zhang2023completionformer}. Note that its affinity map is constructed based on input images, and is jointly optimized with fixed depth patterns. Different from real-world scenarios where various types of depth sensors (e.g., Velodyne LiDAR~\cite{schwarz2010mapping}, Microsoft Kinect~\cite{zhang2012microsoft}, Intel RealSense~\cite{keselman2017intel}, and Apple LiDAR sensor~\cite{luetzenburg2021evaluation}, etc.) are used, the mainstream of standard benchmarks~\cite{kitti_DC_benchmark} for this task is to only use KITTI dataset~\cite{kitti} captured from a 64-Line LiDAR and random samples from Kinect depth camera in NYUv2 dataset~\cite{NYUv2}.

In this work, our primary goal is to build a sensor-agnostic depth estimation model that faithfully works on the various active depth sensors. Inspired by pioneer works in visual prompt methods like SAM~\cite{kirillov2023segment}, we design a novel depth prompt module used with pre-trained models for monocular depth estimation. The depth prompt first encodes the sparse depth information and then fuses it with image features to construct a pixel-wise affinity. A final refinement process is performed with both the affinity and an initial depth map from the pre-trained depth models. To take full advantage of pre-trained models, we conduct a bias tuning~\cite{cai2020tinytl}, which is a well-known memory-efficient technique when applied to pre-trained models. Our proposed method is fine-tuned for only 0.1\% parameters of the models while keeping other parameters frozen.

Our key idea is to reinterpret prompt learning as spatial propagation. We aim to achieve an adaptive affinity from both the depth prompt and the knowledge of the pre-trained monocular depth model~\cite{li2023depthformer}. We demonstrate that the proposed method is generalized well to any sensor type, and it can be extended for various depth foundation models such as \cite{birkl2023midas,spencer2023kick}, which are trained with large-scale datasets for monocular depth estimation to achieve relative scale prediction and zero-shot generalization. To do this, we utilize a variety of public datasets captured from off-the-shelf depth sensors and take real-world scenes by ourselves where the wide depth ranges and structures exist. Additionally, we conduct an extensive ablation study to verify the influence of each component within our framework and evaluate our methodology in a variety of real-world settings. This testing includes zero and few-shot inference exercises across different sensor types, further validating the robustness and adaptability of our proposed solution.

\section{Related Works}

\subsection{Depth Estimation with Sparse Measurement}

Accurate dense depth acquisition requires time-consuming and costly processes~\cite{li2019aads}. As a compromise between inference time and cost, sparse depth measurements based on active sensing manners have been considered as mainstream approaches. Leveraging the sparse measurements and its corresponding images, recent learning-based methods~\cite{LearningAffinitySPN,cspn,dyspn,nlspn,park2023learning,zhang2023completionformer} have been proposed to make dense predictions which enhance the depth resolutions same with the image resolutions via a spatial propagation process. However, due to the dependency on the specific sparse depth pattern and density according to the input devices, such models face challenges in real-world scenarios such as sensor blackouts~\cite{vidacs2021winning}, multipath interference~\cite{bhandari2014resolving,pare2020multipath}, and non-Lambertian surfaces~\cite{tykkala2011direct}, resulting in much fewer sample measurements.

Several studies have explored depth reconstruction from unevenly distributed and sparse input data~\cite{guizilini2021sparse,conti2023sparsity,yin2022towards}. However, they suffer from an issue on a range bias, which provides only limited scan ranges in training datasets. Works in \cite{xia2020generating,dexheimer2023learning} focus on handling extremely sparse conditions (less than $0.1\%$ over its input image); however, they fail to show the generalized performances on scenarios given relatively dense initial depth inputs. To address both issues, we adopt prompt engineering to achieve the model generalization, which has proven its powerful capacity by taking advantage of pre-trained models on downstream tasks, yet remains unsolved for depth-relevant tasks so far~\cite{kirillov2023segment,brown2020language,gu2023systematic}.

\subsection{Prompt Engineering}

Prompt engineering refers to designing specific templates that guide a model to complete missing information in a structured format (e.g., cloze set~\cite{brown2020language}), or generate a valid response along with given input~(e.g., promptable segmentation~\cite{kirillov2023segment}). With the recent success of the large language models~\cite{sarzynska2021detecting,devlin2018bert}, works in \cite{brown2020language,dong2022survey} demonstrate how various natural language processing (NLP) tasks can be reformulated to an incontext learning problem given a pre-defined prompt, which is a useful tool for solving the tasks and benchmarks~\cite{radford2018improving,radford2019language}. 

The advent of prompt engineering has been transformative, with some studies~\cite{liu2023explicit,kirillov2023segment,radford2021learning} extending its application to the computer vision field. Here, it is used to achieve a zero-shot generalization, enabling models to understand new visual concepts and data distributions that they are not explicitly trained on.
The most\,relevant\,work~\cite{kirillov2023segment}\,to\,this\,paper\,designs\,a promptable segmentation model. They construct a prompt encoder to represent user-defined points or boxes with a positional embedding~\cite{tancik2020fourier}. We\,note\,that\,it\,is\,not a\,pixel-wise regression problem which is different from\,our\,task.

\subsection{Foundation Model for Dense Prediction}
Foundation models are designed to be adaptable for various downstream tasks by pretraining on broad data at scale~\cite{bommasani2021opportunities}. In NLP field revolutionized by large-scale models such as GPT series~\cite{brown2020language,radford2018improving,radford2019language}, foundation models in the computer vision field have been becoming popular. Recent advancements in the large-scale foundation models with web-scale image databases have made significant breakthroughs, particularly involving image-to-text correspondences~\cite{jia2021scaling,radford2021learning,yuan2021florence,li2023blip}. These developments have paved the way for more efficient transfer learning~\cite{yuan2021florence,radford2021learning,Alayrac2022Flamingo} and make the zero-shot capabilities better~\cite{jia2021scaling,radford2021learning,li2023blip}. 

Despite these advancements, the foundation models are primarily used in high-level vision tasks such as image recognition~\cite{jia2021scaling,yuan2021florence}, image captioning~\cite{yuan2021florence,Alayrac2022Flamingo,li2023blip}, and text-to-image generation~\cite{li2023blip}. When it comes to low-level vision tasks like depth predictions, these models do not seem to be suitable due to a lack of extensive image/depth data on a metric scale~\cite{liu2023degae,li2023lightweight}. To be specific, a web-scale dataset collection is infeasible because ground truth-level metric scale depths can be obtained only from sensor fusion manners~\cite{richter2015robust}. Although some works~\cite{spencer2023kick,birkl2023midas} have led to the creation of large and diverse datasets for monocular depth prediction, transferring the learned knowledge into other domains remains unexplored. To achieve the sensor-agnostic depth estimation regardless of scene configuration, we take fully advantage of the knowledge from the depth foundation model.

\section{Sensor-agnostic Depth Estimation}
In this section, we start by discussing the three biases which hinder sensor-agnostic depth prediction~(\cref{subsec:Sensor Biases in Depth Estimation}). We then introduce the proposed depth prompt module. Here, we recast the depth prompt design as learning an adaptive affinity construction in spatial propagation for various types of sparse input measurements~(\cref{subsec:Depth Prompting}). Lastly, we provide implementation details of the proposed module~(\cref{subsec:Implementation}).

\subsection{Sensor Biases in Depth Estimation}
\label{subsec:Sensor Biases in Depth Estimation}
Bias issues make learning-based models for visual perceptions hard to achieve their generality~\cite{Nam2021Reducing,Bae2023Digging}. There have been attempts to address the bias problems in image restoration~\cite{Zhang2022Plug}, recognition~\cite{Wang2020Towards} and generation~\cite{Gal2022StyleGANNADA}. Among them, the sensor-bias issue~\cite{Dijk2019How} is also considered as one of the crucial research topics. In particular, since a variety of depth sensors types is available, there is no generalization\,method\,to\,cover every depth sensor types, while the solutions to the same type (e.g., different LiDAR configurations~\cite{yin2022towards}) exist. In this part, we empirically investigate 3-sensor biases, e.g., \textit{sparsity}, \textit{pattern}, and \textit{range bias} before an\,introduction\,to\,our\,solution. 

\begin{figure}[t]
\centering
 \includegraphics[width=1.0\columnwidth]{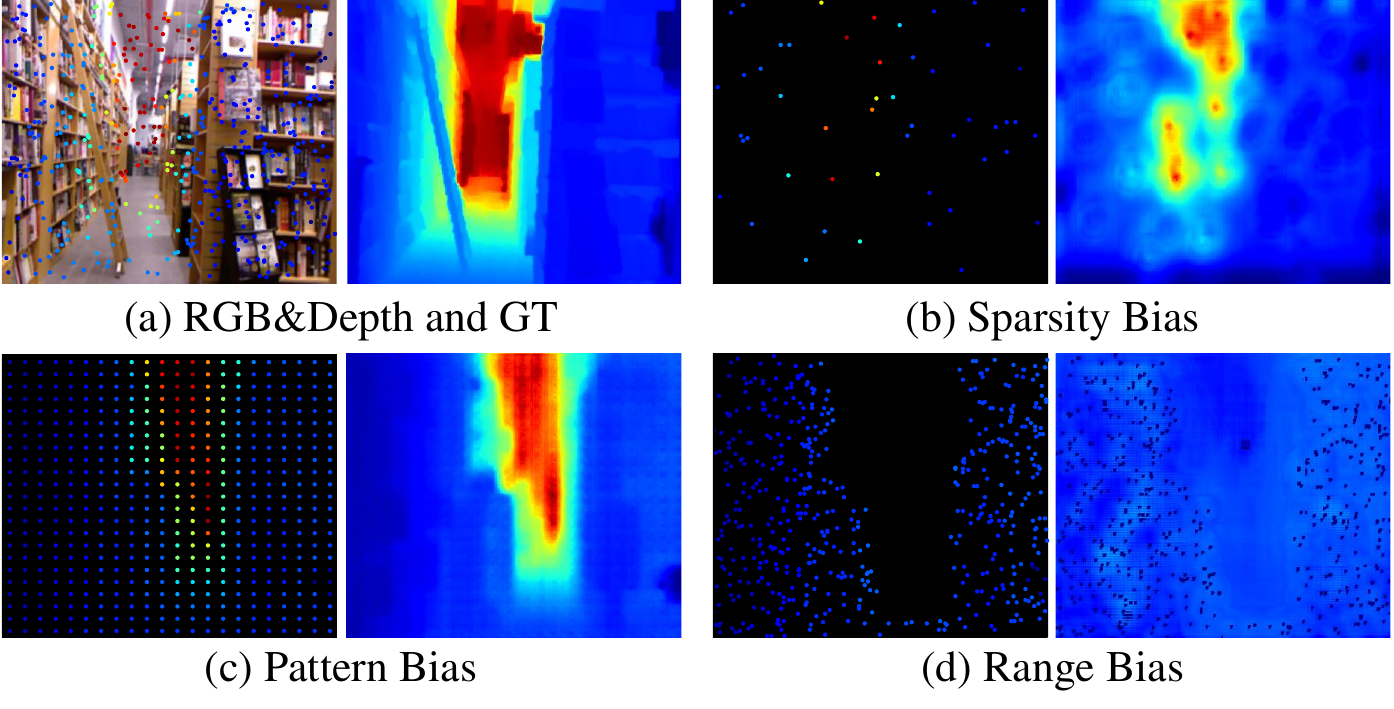}
 \vspace{-7.5mm}
 \caption{Examples of sensor biases. Depth estimation with an active sensor suffers from bias problems, including fixed density and pattern, and inherent scan range of sensors used.}
 \label{fig:bias}
\end{figure}

Firstly, if a learning-based model is trained on data with a certain density (e.g., 500 random samples in training), it will suffer from \textit{sparsity bias}, which makes high-fidelity depth maps difficult if fewer samples are available in the test phase, as shown in \cref{fig:bias}-(b). The sparser sample measurements are also common due to sensor blackout, occlusion, or changing environmental conditions. This has hampered the practical utility of learning-based depth estimation in real-world scenarios. Second, \textit{pattern bias} shows the performance degradation if depth patterns vary between training and test phases, even with the same number of depth points. When we intentionally shift the input depth pattern in the inference, it indicates that the existing model is biased toward the fixed location of input depth points in \cref{fig:bias}-(c). This makes a unified depth prediction model difficult to be applied to other sensor types. Lastly, \textit{range bias}, arising when attempted to take scene structures beyond the limited scan range of the sensor, also prevents sensor-agnostic depth estimation as shown in \cref{fig:bias}-(d).

\subsection{Depth Prompting}
\label{subsec:Depth Prompting}
\begin{figure}[t]
\centering
 \includegraphics[width=1.\columnwidth]{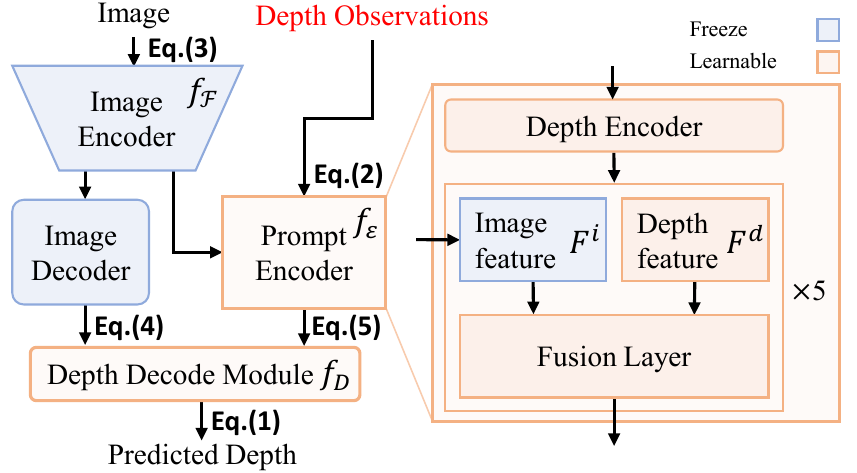}
 \vspace{-7.5mm}
 \caption{An overview of the proposed architecture. We design a depth prompt module to construct an adaptive affinity map $A_{ada}$, which guides the propagation of given depth information. }
 \label{fig:method}
\end{figure}

To realize sensor-agnostic depth estimation without any sensor bias, we take an inspiration from prompt learning in NLP, which designs a specific template to guide a model for a valid response along with a given input~\cite{brown2020language,dong2022survey}. We aim to design a prompt module for depth modality by defining a unified embedding space to represent learned features from any type of input measurements~(\cref{fig:method}). Here, we use an input depth map as a template for our depth prompt module, and the sensor-agnostic depth prediction is achieved by fusing the template, features\,from\,the\,embedding\,space,\,and\,image\,features.

\noindent\textbf{Revisiting Spatial Propagation.}\quad In this work, our key idea is to reinterpret the depth prompt design as spatial propagation, which predicts dense depth maps from input sparse measurements guided by image-dependent affinity weights. 
We formulate the conventional spatial propagation process: 
\begin{equation}
    \!\!D^{t+1}_{(x,y)}\!=\!A(x,y) \odot D^{0}_{(x,y)} {+} \!\!\!\!\!\! \!\!\!\! \sum_{(l,m) \in \mathcal{N}_{(x,y)}} \!\!\!\!\!\! \!\!\! A(l,m) \odot D^{t}_{(l,m)},
\label{eqn:spatialpropagation}
\end{equation}
where ${D^t_{(x,y)}} \in \mathbb{R}^{1 \times H \times W}$ refers to a depth map for each propagation step $t$. $(x,y)$ and $H,W$  means a spatial coordinate and the height and width of an input image, respectively.
$D^{0}_{(x,y)}$ and $A$ indicate an initial depth and a pixel-wise affinity map, respectively. $\odot$ operator denotes an element-wise product. $(l,m) \in \mathcal{N}_{(x,y)}$ refers to 8-directional neighboring pixels over the reference pixel $(x,y)$. 

Even in the same scene, the affinity map $A$ can vary according to the input depth type which is dependent on sensors used. That's, the affinity map should be adaptive to various input changes. However, affinity maps from previous spatial propagation methods~\cite{LearningAffinitySPN,cspn,dyspn,nlspn,park2023learning,zhang2023completionformer} are invariant because they are learned from a certain type of input depths. We address this issue by designing a depth encoder to learn features for a diverse set of sensors and by projecting them into the unified embedding space.

\noindent\textbf{Depth Feature Extraction.}\quad To do this, we adopt an encoder-decoder structure to efficiently encode both positional and sparsity information of an input depth map. The encoder takes a depth map as an input, and then the decoder constructs an affinity map with the same size of the depth map. After that, the prompt embedding is combined with image features to bring\,boundary\,and\,context\,information.

To be specific, we use ResNet34~\cite{he2016deep} to extract the depth features~\cite{Zhang_2023_ICCV,Qiu_2019_CVPR,lu2020depth}. We downsample the features by $1/2$, $1/4$, $1/8$, $1/16$, and $1/32$, and then feed them into the decoder with skip connections. Given a sparse depth $D_S$ , our depth prompt encoder $f_{\mathcal{E}}$ yields both a prompt embedding $F^d$ and multi-scale features $F^d_k$ as below:
\begin{equation}
F^d, F^d_k = f_{\mathcal{E}} (D_S),
\label{eqn:encoder}
\end{equation}
where $k$ is an index of the downsampled features.


\noindent\textbf{Depth Foundation Model.}\quad 
%
%
Until now, large-scale depth models~\cite {spencer2023kick,birkl2023midas,li2023depthformer} have been tailored to monocular depth estimation. Leveraging a large-scale monocular depth dataset, the models are able to provide relative depth maps, which is the only option as a foundation model.

Given a single image $I \in \mathbb{R}^{3 \times H \times W}$, the pre-trained depth model $f_{\mathcal{F}}$ outputs an initial depth map $\hat{D}_{I}$ and multi-scale intermediate features $F^i_k$:
\begin{equation}
\hat{D}_{I}, F^i_k = f_{\mathcal{F}}(I,\Theta_{f_{\mathcal{F}}}),
\label{eqn:foundation}
\end{equation}
where $\Theta_{f_{\mathcal{F}}}$ indicates parameters of the foundation model, which keeps frozen during both training and inference.

Here, we need to effectively transfer the pre-trained knowledge of the foundation models to a range of sensors via prompt engineering.
We adopt a bias tuning~\cite{jia2022visual,cai2020tinytl}, which is more effective for dense prediction tasks than other tuning protocols~\cite{houlsby2019parameter,jia2022visual}. This is used to update bias terms and to freeze the rest of the backbone parameters. As a result, the bias tuning contributes to preserving the high-resolution details and context information acquired in the initial extensive training phase~\cite{cai2020tinytl}, which will be analyzed in~\cref{subsec:Ablation}.

Next, to infer depth maps with absolute scales, we merge the relative depth from the foundation model with the sensor measurements. Due to the nature of spatial propagation, which mainly refines neighboring depth values over given seed points, we cannot obtain proper depth values for regions where no initial depth points are available. To address this limitation, we perform an additional processing with a least-square solver~\cite{marquardt1963algorithm} to produce the consistent $D_I$ when applied to other sensor types. 
The process uses both an initial depth from the foundation model and a sparse depth $D_S$ in order to perform a global refinement. By solving the least square equation, we can obtain the solution $p \in \mathbb{R}$ as below:
\begin{equation}
    \hat{p} = \min_{p} || p\hat{D}_{I}^V  - D_S ||_F,    
    \label{eq:lsq}
\end{equation}
where $||\cdot||_F$ denotes the Frobenius norm, and ${D}_{I}^V \subset \hat{D}_{I}$ refers to a set of pixels corresponding to valid depth points $D_S$.
The solution $\hat{p}$ is multiplied with the initial depth $\hat{D}_{I}$, whose result becomes $D^0_{(x,y)}$ of \cref{eqn:spatialpropagation}. 
Since this pre-calculation makes an initial depth for the spatial propagation in \cref{eqn:spatialpropagation} better, we can cover larger unknown areas than those without \cref{eq:lsq}.




\noindent\textbf{Decoder for Adaptive Affinity.}\quad A decoder in our prompt module reconstructs an affinity map using the image embedding from the foundation model encoder $f_{\mathcal{F}}$~(\cref{eqn:foundation}) and a set of prompt embeddings from the prompt encoder $f_{\mathcal{E}}$~(\cref{eqn:encoder}). We concatenates the prompt embeddings $F^d$, intermediate features $F^d_i$ from the prompt encoder and multi-scale image features $F^i_k$, and then yield $A_{\text{ada}} \in \mathbb{R}^{C^2 \times H \times W}$ where $C$ is a hyper-parameter to define the propagation ranges and set to 7 as below:
\begin{equation}
    A_{\text{ada}} = f_{\mathcal{D}} (F^d,F^d_k,F^i_k).    
    \label{eq:affinity_ada}
\end{equation}

Finally, we substitute the conventional affinity map $A$ in \cref{eqn:spatialpropagation} into our $A_{\text{ada}}$ above.
Thanks to the feature fusion from both the prompt embedding and the foundation model with the bias-tuning, we can successfully decode an affinity map to account for different types of input measurements. In addition, the least square solver in \cref{eq:lsq} allows the spatial propagation to take consistent initial depth maps as input, regardless of the sensor variations. As a result, we achieve the adaptiveness/robustness in the proposed framework.

\subsection{Implementation}
\label{subsec:Implementation}

\noindent\textbf{Random Depth Augmentation.}\quad
For more generality, we adopt random depth augmentation (RDA). We sample depth points from relatively dense depth maps to simulate sparser input depth scenarios. For example, we extract 4-Line depth values from 64-Line depth maps in the KITTI dataset~\cite{kitti}. In addition, we train our framework from general to extreme cases (e.g., from standard 500 random samples to only 1 depth point in the NYUv2 dataset~\cite{NYUv2})



\noindent\textbf{Loss Functions.}\quad 
The proposed framework is trained in a supervised manner with the linear combination of two loss functions: (1) Scale-Invariant (SI) loss~\cite{eigen2014depth} for an initial depth from the depth foundation model $f_{\mathcal{F}}$~(\cref{eqn:foundation}); (2) A combination of $L_1$ and $L_2$ losses for a final dense depth. 

An initial depth map is predicted by minimizing the difference between $\hat{D}_{I}$ and its ground truth depth map $D^{gt}$ for valid pixels ${v \in V}$. Let $\delta_v = \log \hat{D}_{I}(v) - \log D^{gt}(v)$, the SI loss~$L_{\text{SI}}$ is defined as below:
\begin{equation}
L_{\text{SI}}(\hat{D}_{I},D^{gt}) =\frac{1}{|V|} \sum_{v \in V}\left(\delta_v\right)^2-\frac{\lambda}{|V|^2}\left(\sum_{v \in V} \delta_v\right)^2,
\label{eqn:siloss}
\end{equation}
where we set $\lambda=0.85$ in all experiments, following the previous work~\cite{li2023depthformer}.

Next, our framework infers a dense depth $\hat{D}$ in~\cref{eqn:spatialpropagation} based on the valid pixels ${v \in V}$ of its ground truth depth $D^{gt}$ as well. 
For this, we use a loss $L_{\text{comb}}$ based on both $L_{1}$ and $L_{2}$ distances as follows:
\begin{equation}
\begin{split}
&L_{\text{comb}}(\hat{D}, D^{gt})= \\
&\frac{1}{|V|} \sum_{v \in V}\left(\left|\hat{D}(v)-D^{gt}(v)\right| +\left|\hat{D}(v)-D^{gt}(v)\right|^2\right),        
\label{eqn:l1l2loss}
\end{split}
\end{equation}
In total, our framework is optimized by minimizing the final loss $\mathcal{L}$ as below:
\begin{equation}
    \mathcal{L}= L_{\text{comb}}(\hat{D}, D^{gt})+\mu L_{\text{SI}}(\hat{D}_{I},D^{gt}).
\end{equation}
where $\mu$ is a balance term and empirically set to 0.1.

%

 %
%
%

\noindent\textbf{Training Details.}\quad 
We utilize a SoTA monocular depth estimation method, termed DepthFormer~\cite{li2023depthformer}, as a primary backbone to validate our method effectively transfer the knowledge of large-scale depth model into our sensor-agnostic model.
Our framework is implemented in public PyTorch~\cite{paszke2017automatic}, trained for 25 epochs on four RTX 3090TI GPUs using Adam~\cite{kingma2014adam} optimizer, with 228 × 304 and 240 × 1216 input resolution of NYU and KITTI dataset, respectively. Note that we resize the input RGB images to keep their ratio of height and width toward the foundation model used. 
The initial learning rate is $2 \times 10^{-3}$, and then scaled down with coefficients 0.5, 0.1, and 0.05 every 5 epochs after 10$^{\text{th}}$ epoch. The total training process for the NYU dataset takes approximately half a day, with an inference time of 0.06 seconds.  For the KITTI dataset, the training time is about 1.5 days, with an inference time of 0.38 seconds. The framework comprises 53.4 million learnable parameters, which includes 0.1M dedicated to tuning the foundational model.

\begin{table*}[!ht]
\centering
\LARGE
\resizebox{\textwidth}{!}{%
\begin{tabular}{c||ccc|ccc|ccc|ccc|ccc|ccc}
\toprule
\multirow{2}{*}{\# Samples} & \multicolumn{3}{c}{200}                                  & \multicolumn{3}{c}{100}                             & \multicolumn{3}{c}{32}                              & \multicolumn{3}{c}{8}                               & \multicolumn{3}{c}{4}                               & \multicolumn{3}{c}{1}                               \\ \cmidrule{2-19}
                     & RMSE                 & MAE             &\!\!DELTA1\!\!         & RMSE            & MAE             &\!\!DELTA1\!\!         & RMSE            & MAE             &\!\!DELTA1\!\!         & RMSE            & MAE             &\!\!DELTA1\!\!         & RMSE            & MAE             &\!\!DELTA1\!\!         & RMSE            & MAE             &\!\!DELTA1\!\!         \\ \midrule
CSPN                 & 0.1563               & 0.0707          & 0.9868          & 0.2795          & 0.1491          & 0.9585          & 0.6306          & 0.4344          & 0.7310          & 0.9688          & 0.7417          & 0.5031          & 1.0399          & 0.8186          & 0.4473          & 1.1093          & 0.8850          & 0.4108          \\
S2D                  & 0.1871               & 0.1031          & 0.9829          & 0.2733          & 0.1611          & 0.9598          & 0.4084          & 0.2615          & 0.9025          & 1.0982          & 0.8236          & 0.4537          & 1.7385          & 1.4294          & 0.1787          & 1.8446          & 1.5477          & 0.1416          \\
NLSPN\dag               & 0.1358               & \textbf{0.0553} & \textbf{0.9899} & 0.2452          & 0.1125          & 0.9693          & 0.5541          & 0.3427          & 0.8254          & 0.9564          & 0.7023          & 0.5479          & 1.0775          & 0.8273          & 0.4511          & 1.1929          & 0.9521          & 0.3616          \\
DySPN                & 0.1532               & 0.0686          & 0.9880          & 0.3174          & 0.1799          & 0.9345          & 0.6603          & 0.4838          & 0.6751          & 0.9635          & 0.7586          & 0.4913          & 1.0351          & 0.8304          & 0.4484          & 1.1079          & 0.9014          & 0.4110          \\
CostDCNet\dag            & 0.1455               & 0.0606          & 0.9887          & 0.2809          & 0.1300          & 0.9592          & 0.6887          & 0.4144          & 0.7735          & 1.1685          & 0.8479          & 0.4864          & 1.3097          & 0.9973          & 0.3915          & 1.4255          & 1.1297          & 0.3065          \\
CompletionFormer\dag    & \textbf{0.1352}      & 0.0583          & 0.9898          & 0.3553          & 0.2125          & 0.9069          & 0.7921          & 0.6160          & 0.5250          & 1.0647          & 0.8536          & 0.3830          & 1.1259          & 0.9112          & 0.3526          & 1.2091          & 0.9878          & 0.3167          \\
Ours                 & 0.1435               & 0.0642          & 0.9881          & \textbf{0.1778} & \textbf{0.0870} & \textbf{0.9812} & \textbf{0.2472} & \textbf{0.1452} & \textbf{0.9546} & \textbf{0.3673} & \textbf{0.2464} & \textbf{0.9133} & \textbf{0.3827} & \textbf{0.2627} & \textbf{0.9031} & \textbf{0.4040} & \textbf{0.2848} & \textbf{0.8935} \\ \bottomrule
\end{tabular}%
}
\vspace{-3.8mm}
\caption{Quantitative Results on NYUv2. \dag~indicates that the publicly available code and model weight is utilized in this experiment.}
\vspace{-3.5mm}
\label{tab:nyu}
\end{table*}

\begin{table*}[!ht]
\LARGE
\centering
\resizebox{\textwidth}{!}{%
\begin{tabular}{c||ccc|ccc|ccc|ccc|ccc|ccc}
\toprule
\multirow{3}{*}{\# Lines} & \multicolumn{3}{c}{32}                                                               & \multicolumn{3}{c}{16}                                                               & \multicolumn{3}{c}{8}                                                                & \multicolumn{3}{c}{4}                                                                & \multicolumn{3}{c}{2}                                                                 & \multicolumn{3}{c}{1}                                                                 \\ \cmidrule{2-19}
                          & RMSE                       & MAE                        &\!\!DELTA1\!\!                    & RMSE                       & MAE                        &\!\!DELTA1\!\!                    & RMSE                       & MAE                        &\!\!DELTA1\!\!                    & RMSE                       & MAE                        &\!\!DELTA1\!\!                    & RMSE                        & MAE                        &\!\!DELTA1\!\!                    & RMSE                        & MAE                        &\!\!DELTA1\!\!                    \\ \midrule
CSPN                      & 1.3644                     & 0.4008                     & 0.9876                     & 2.0935                     & 0.7532                     & 0.9596                     & 3.5806                     & 1.6679                     & 0.8580                     & 5.7438                     & 3.1650                     & {0.6656} & 8.9143                      & 5.6544                     & 0.4478                     & 12.5666                     & 8.2679                     & 0.3379                     \\
S2D                       & {1.8133} & {0.7426} & {0.9623} & {2.6886} & {1.1670} & {0.9232} & {4.5806} & {2.6697} & {0.6727} & {7.2951} & {4.9223} & {0.3866} & {10.1624} & {6.8438} & {0.2854} & {12.2972} & {7.9316} & {0.2057} \\
NLSPN\dag                     & 1.1894                     & 0.3536                     & 0.9923                     & 1.9279                     & 0.6976                     & 0.9675                     & 3.2285                     & 1.5482                     & 0.8692                     & 4.7571                     & 2.5976                     & 0.7267                     & 6.0305                      & 3.8779                     & 0.4904                     & 8.8244                      & 5.2859                     & 0.3949                     \\
DySPN                     & 1.6758 & 0.5449 & 0.9871 & 2.3979 & 0.9096 & 0.9624 & 3.4687 & 1.5774 & 0.883 & 5.2374 & 2.8549 & 0.6981 & 6.5413 & 4.0182 & 0.5118 & 9.5199 & 5.3260 & 0.4637  \\
CompletionFormer\dag          & 1.2513                     & 0.3844                     & 0.9912                     & 2.1857                     & 0.8403                     & 0.9627                     & 3.6505                     & 1.7687                     & 0.8577                     & 6.2532                     & 3.4800                     & 0.6787                     & 8.9682                      & 5.9899                     & 0.4672                     & 12.7693                     & 9.0019                     & 0.3414                     \\ 
SAN\dag  & 1.8188                     & 0.8160                     & 0.9793                     & 2.8866                     & 1.5915                     & 0.9087                     & 3.7936                     & 1.8339                     & 0.8967                     & 4.5894                     & 2.3092                     & 0.8523                     & 4.1416                      & 2.1280                     & 0.8591                     & 4.4444                      & 2.3270                     & 0.8153                     \\ \midrule
Ours                      & \textbf{1.1465}            & \textbf{0.3472}            & \textbf{0.9942}            & \textbf{1.3512}            & \textbf{0.4134}            & \textbf{0.9922}            & \textbf{1.6419}            & \textbf{0.5470}            & \textbf{0.9888}            & \textbf{1.9507}            & \textbf{0.7629}            & \textbf{0.9809}            & \textbf{2.3841}             & \textbf{1.1976}            & \textbf{0.9505}            & \textbf{2.8234}             & \textbf{1.2678}            & \textbf{0.9535}            \\ \bottomrule
\end{tabular}%
}
\vspace{-3.8mm}
\caption{Quantitative Results on KITTI DC. \dag~indicates that the publicly available code and weight are used in this experiment.}
\vspace{-1mm}
\label{tab:kitti}
\end{table*}
\begin{table}[t]
\LARGE
\centering
\resizebox{\columnwidth}{!}{%
\begin{tabular}{c||cccc|cccc}
\toprule 
\multirow{3}{*}{\begin{tabular}[c]{@{}c@{}}KITTI\\ $\downarrow$ \\ NYU\end{tabular}}  & \multicolumn{4}{c}{100-shot}  & \multicolumn{4}{c}{10-shot}                                       \\ \cmidrule{2-9}
                                         & \multicolumn{2}{c}{500}           & \multicolumn{2}{c}{50}            & \multicolumn{2}{c}{500}           & \multicolumn{2}{c}{50}            \\ \cmidrule{2-9}
                     & RMSE            & MAE             & RMSE            & MAE             & RMSE            & MAE             & RMSE            & MAE             \\ \midrule
CSPN                                     & 0.1848          & 0.0909          & 1.0235          & 0.7220          & 0.6741          & 0.5038          & 1.2859          & 1.0251          \\
NLSPN                                    & 0.1940           & 0.1075          & 0.5917          & 0.4285          & 0.7367          & 0.5682          & 1.2798          & 1.0378          \\ 
CompletionFormer & 0.2259 & 0.1287  & 1.1560 & 0.4589 & 0.5881 & 0.8758 & 1.3870 & 1.1430 \\ \midrule
Ours                                     & \textbf{0.1748} & \textbf{0.0796} & \textbf{0.4697} & \textbf{0.3048} & \textbf{0.5466} & \textbf{0.4284} & \textbf{0.7990} & \textbf{0.6482} \\ \bottomrule 
\multirow{3}{*}{\begin{tabular}[c]{@{}c@{}}NYU \\ $\downarrow$ \\ KITTI\end{tabular}}   & \multicolumn{4}{c}{100-shot}                   & \multicolumn{4}{c}{10-shot}                       \\ \cmidrule{2-9}
                                         & \multicolumn{2}{c}{64}            & \multicolumn{2}{c}{8}             & \multicolumn{2}{c}{64}            & \multicolumn{2}{c}{8}             \\ \cmidrule{2-9}
                                          & RMSE            & MAE             & RMSE            & MAE             & RMSE            & MAE             & RMSE            & MAE             \\ \midrule
CSPN                                     & 1.2803          & 0.3645          & 5.8021          & 3.4752          & 2.2717          & 0.9287          & 9.9108          & 6.1467          \\
NLSPN                                    & 1.5156          & 0.5194          & 4.5913          & 2.7022          & 3.6842          & 1.8959          & 10.8715         & 5.7224          \\ 
CompletionFormer                         & 1.3404           & 0.3770         & 5.0787          & 3.2564 & 2.3280 & 1.0014& 10.8328 & 7.0782  \\ \midrule
Ours                                     & \textbf{1.2752} & \textbf{0.3534} & \textbf{4.5874} & \textbf{2.4834} & \textbf{2.1108} & \textbf{0.7846} & \textbf{5.6511} & \textbf{3.1699} \\ \bottomrule 
\end{tabular}%
}
\vspace{-3.8mm}
\caption{Cross-validation between Indoor and Outdoor dataset.}
\label{tab:nyu_kitti}
\vspace{-4mm}
\end{table}

\section{Experiment}
\label{sec:Experiment}
In this section, we conduct comprehensive experiments to evaluate the impact of our depth prompting module on sensor-agnostic depth estimation. First, we briefly describe the experimental setup~(\cref{subsec:Experiment Setup}), and present comparative results against various state-of-the-art (SoTA) methods on standard benchmark datasets~(\cref{subsec :Quantitative results}).
Moreover, we provide an in-depth examination of bias issues in sensor~(\cref{subsec:SPR}) as well as an ablation study to demonstrate the effect of each component in our method~(\cref{subsec:Ablation}). Lastly, we offer qualitative results to show zero generalization of our method on commercial sensors~(\cref{subsec: Zero-shot Inference on Commercial Sensors}).




\subsection{Experiment Setup}
\label{subsec:Experiment Setup}

\noindent\textbf{Evaluation Protocols.}\quad
For our comparative experiments, we select a range of SoTA methods for depth estimation from sparse measurements. These include a series of spatial propagation networks such as CSPN~\cite{cspn}, S2D~\cite{ma2018self}, NLSPN~\cite{nlspn}, DySPN~\cite{dyspn}, CostDCNet~\cite{kam2022costdcnet}, and CompletionFormer~\cite{zhang2023completionformer}. Additionally, we choose SAN~\cite{guizilini2021sparse}, which are designed to adapt various sparse setups. We use common quantitative measures of depth quality: root mean square error (RMSE, unit: meter), mean absolute error (MAE, unit: meter), and inlier ratio~(DELTA1, $\delta<1.25$).

\noindent\textbf{Datsets: NYUv2 and KITTI DC.}\quad
We utilize the NYU Depth V2 dataset, an indoor collection featuring 464 scenes captured with a Kinect sensor. Following the official train/test split, we use 249 scenes (about 50K samples) in training phase, and 215 scenes (654 samples) are tested for the evaluation. The NYU Depth V2 dataset provides 320×240 resolutions. We use the center-cropped image whose resolution is 304×228 and randomly sample 500 points to simulate the sparse depth. 

For outdoor scenarios, we choose a KITTI DC~\cite{kitti_DC_benchmark} dataset with 90K samples. Each sample includes color images and aligned sparse depth data (about 5\% density over image resolution) captured using a high-end Velodyne HDL-64E LiDAR sensor. The images have 1216×352 resolution. The dataset is divided into training (86K samples), validation (7K samples), and testing segments (1K samples). Ground truth is established by accumulating multiple LiDAR frames and filtering out errors, which results in denser LiDAR depths (about 20\% density).


%
%

\subsection{Experimental Results}
\label{subsec :Quantitative results}


\noindent\textbf{Sensor Agnsoticity.}\quad
We assess the versatility of our method and the SoTA methods across various density levels. They are commonly trained with a standard training protocol, e.g., 500 random depth samples from the NYUv2 dataset and 64 lines on the KITTI DC dataset. We test them under exactly the same conditions. For the NYUv2 dataset, we sample fewer samples (from 200 to 1 depth point) than that used in training phase. In addition, we use less scanning lines (from 32 to 1 line) than the original KITTI dataset. 

As shown in~\cref{tab:nyu,tab:kitti}, our method consistently provides the superior results in almost test conditions. While methods such as NLSPN~\cite{nlspn} and CompletionFormer~\cite{zhang2023completionformer} demonstrate their robustness with 200 samples in NYUv2 dataset and the 32-line scenario in KITTI dataset, respectively, our approach outperform the SoTA models in the more challenging scenarios. The depth prompt encoder contributes to constructing an adaptive representation for randomly given seeds, regardless of the pattern and density. 


We observe that a majority of SoTA methods heavily depend on predefined input configurations. Spatial propagation, a prevalent technique among these methods, relies on relations among neighboring pixels, requiring a substantial number of seeds to cover an entire scene. This dependency results in significant performance deterioration in scenarios with sparser initial depth seeds. 
In addition, SAN~\cite{guizilini2021sparse}, which are engineered to merge depth and image features at the late fusion to achieve stability in varying sparsity conditions, also encounter the performance drop in \cref{tab:nyu,tab:kitti}.

In contrast, thanks to the knowledge of the foundation model and the depth-oriented prompt engineering, our method achieves relatively stable performance. Our prompt module enables the construction of an adaptive affinity map according to the distribution of input data, whose effectiveness is enlarged by the zero-shot generalization for unseen visual attributes and data distributions.

\noindent\textbf{Cross-validation between Indoor and Outdoor.}\quad
To conduct a cross-validation between outdoor and indoor scenarios, we finetune our model and the comparison methods with only 10 and 100 images. Since active sensors provide metric depth to the model, the domain adaption via a few ground-truth level annotations is inevitable. Following~\cite{wei2023imitation}, we randomly select the pair of images and depth data. As shown in \cref{tab:nyu_kitti}, our method shows the superior performance than the SoTA methods, which demonstrates the model's effectiveness in preserving visual features across varying scan ranges and its successful adaptation of the pre-trained knowledge to different domains.




\begin{table*}[t]
\LARGE
\centering
\resizebox{\textwidth}{!}{%
\begin{tabular}{c||ccc|ccc|ccc|ccc|ccc|ccc}
\toprule
\multirow{2}{*}{} & \multicolumn{3}{c}{\begin{tabular}[c]{@{}c@{}}Sparsity \\ (500 \!$\rightarrow$\! 50)\end{tabular}} & \multicolumn{3}{c}{\begin{tabular}[c]{@{}c@{}}Sparsity Rev.\\ (50 \!$\rightarrow$\! 500)\end{tabular}} & \multicolumn{3}{c}{\begin{tabular}[c]{@{}c@{}}Pattern\\ (Random \!$\rightarrow$\! \text{Grid})\end{tabular}} & \multicolumn{3}{c}{\begin{tabular}[c]{@{}c@{}}Pattern Rev.\\ (Grid \!$\rightarrow$\! \text{Random})\end{tabular}} & \multicolumn{3}{c}{\begin{tabular}[c]{@{}c@{}}Range \\ (3m$\sim$10m \!$\rightarrow$\! \text{0m$\sim$3m})\end{tabular}} & \multicolumn{3}{c}{\begin{tabular}[c]{@{}c@{}}Range Rev.\\ (0m$\sim$3m \!$\rightarrow$\! \text{3m$\sim$10m})\end{tabular}} \\ \cmidrule{2-19}
                  & RMSE            & MAE             &\!\!DELTA1\!\!         & RMSE              & MAE               &\!\!DELTA1\!\!           & RMSE              & MAE               &\!\!DELTA1\!\!           & RMSE                & MAE                &\!\!DELTA1\!\!            & RMSE                & MAE                &\!\!DELTA1\!\!            & RMSE                & MAE                  &\!\!DELTA1\!\!             \\ \midrule
CSPN              & 0.4902          & 0.3102          & 0.8323          & 0.1538            & 0.0802            & 0.9877            & 0.1108            & 0.0468           & 0.9937            & 0.7657              & 0.5401             & 0.6310              & 0.3419              & 0.2235             & 0.8055             & 0.2869              & 0.1533               & 0.9466              \\
S2D               & 1.3680           & 1.0730           & 0.3006          & 0.5608            & 0.3134            & 0.8827            & 0.4794            & 0.4331            & 0.6836            & 0.8988              & 0.6230              & 0.6265             & 0.8430               & 0.6048             & 0.6208             & 0.8322              & 0.5514               & 0.7678              \\
NLSPN             & 0.4639          & 0.3018          & 0.8554          & 0.1516            & 0.0758            & 0.9882            & 0.1136            & 0.0466            & 0.9933            & 1.2200                & 0.9202             & 0.3982             & 0.3758              & 0.2553             & 0.7871             & 0.3753              & 0.2329               & 0.9207              \\
DySPN             & 0.4473          & 0.2700            & 0.8832          & 0.1487            & 0.0779            & \textbf{0.9887}   & 0.1088            & 0.0439            & 0.9935            & 0.6700                & 0.4117             & 0.7461             & 0.3908              & 0.2654             & 0.7693             & 0.3891              & 0.2138               & 0.9237              \\
CostDCNet         & 0.4701          & 0.2946          & 0.8569          & 0.1458            & 0.0717            & 0.9883            & 0.1248            & 0.0557            & 0.9921            & 0.4164              & 0.2649             & 0.8955             & 0.2160               & 0.1265             & 0.9449             & 0.2205              & 0.0992               & \textbf{0.9788}     \\
CompletionFormer  & 0.4776          & 0.2957          & 0.8510           & 0.1486            & 0.0754            & 0.9879            & 0.1183            & 0.0476            & 0.9925            & 0.8862              & 0.5993             & 0.6276             & 0.3486              & 0.2347             & 0.8207             & 0.6187              & 0.3713               & 0.8614              \\ \midrule
Ours+MiDaS        & 0.4472          & 0.2787          & 0.8567          & 0.1722            & 0.0754            & 0.9827            & 0.1403            & 0.0549            & 0.9884            & 0.4478              & 0.2840              & 0.8608             & 0.2334              & 0.1257             & 0.9691             & 0.2803              & 0.1252               & 0.9574              \\
Ours+KBR          & \textbf{0.3632} & \textbf{0.2282} & \textbf{0.8939} & 0.1503            & 0.0651            & 0.9865            & 0.1170             & 0.0449            & 0.9922            & 0.3133              & 0.1980              & 0.9434             & \textbf{0.2007}     & \textbf{0.1024}    & 0.9697             & \textbf{0.2110}      & \textbf{0.0945}      & 0.9776              \\
Ours  & 0.3997          & 0.2418          & 0.8825          & \textbf{0.1453}   & \textbf{0.0634}   & 0.9874            & \textbf{0.1081}   & \textbf{0.0419}   & \textbf{0.9937}   & \textbf{0.2961}     & \textbf{0.1766}    & \textbf{0.9291}    & 0.2060               & 0.1075             & \textbf{0.9701}    & 0.2328              & 0.0958               & 0.9693 \\ \bottomrule
\end{tabular}%
}
\vspace{-3.8mm}
\caption{Case study on \textit{sparsity}, \textit{pattern}, and \textit{range} biases on the NYUv2 dataset.}
\label{tab:SPR_NYU}
\vspace{-4mm}
\end{table*}

\begin{table*}[h]
\LARGE
\centering
\resizebox{\textwidth}{!}{%
\begin{tabular}{c||ccc|ccc|ccc|ccc|ccc|ccc}
\toprule
\multirow{2}{*}{} & \multicolumn{3}{c}{\begin{tabular}[c]{@{}c@{}}Sparsity \\ (64 \!$\rightarrow$\! 8)\end{tabular}} & \multicolumn{3}{c}{\begin{tabular}[c]{@{}c@{}}Sparsity Rev.\\ (8 \!$\rightarrow$\! 64)\end{tabular}} & \multicolumn{3}{c}{\begin{tabular}[c]{@{}c@{}}Pattern\\ (Line \!$\rightarrow$\! \text{Random})\end{tabular}} & \multicolumn{3}{c}{\begin{tabular}[c]{@{}c@{}}Pattern Rev.\\ (Random \!$\rightarrow$\! \text{Line})\end{tabular}} & \multicolumn{3}{c}{\begin{tabular}[c]{@{}c@{}}Range \\ (15m$\sim$80m \!$\rightarrow$\! \text{0m$\sim$15m})\end{tabular}} & \multicolumn{3}{c}{\begin{tabular}[c]{@{}c@{}}Range Rev.\\ (0m$\sim$15m \!$\rightarrow$\! \text{15m$\sim$80m})\end{tabular}} \\ \cmidrule{2-19}
                  & RMSE            & MAE             &\!\!DELTA1\!\!         & RMSE              & MAE               &\!\!DELTA1\!\!           & RMSE              & MAE               &\!\!DELTA1\!\!           & RMSE                & MAE                &\!\!DELTA1\!\!            & RMSE                & MAE                &\!\!DELTA1\!\!            & RMSE                & MAE                  &\!\!DELTA1\!\!             \\ \midrule
CSPN             & 3.943          & 1.986          & 0.827          & 1.256          & 0.373          & 0.989          & 0.798          & 0.304          & 0.998          & 1.635          & 0.380          & 0.985          & 11.283         & 4.978          & 0.726          & 8.949          & 6.751          & 0.415          \\
S2D              & 4.814          & 2.918          & 0.682          & 2.721          & 1.352          & 0.963          & 2.007          & 1.467          & 0.835          & 3.222          & 1.788          & 0.921          & 11.438         & 6.499          & 0.406          & 11.418         & 8.918          & 0.300          \\
NLSPN            & 5.052          & 2.771          & 0.774          & 1.539          & 0.718          & 0.986          & 1.116          & 0.553          & 0.995          & 1.889          & 0.412          & 0.983          & 11.857         & 7.580          & 0.204          & 13.727         & 11.186         & 0.221          \\
DySPN            & 4.106          & 2.136          & 0.786          & 1.269          & 0.394          & 0.991          & 0.837          & 0.317          & 0.997          & 1.856          & 0.489          & 0.982          & 11.127         & 5.028          & 0.700          & 9.969          & 7.641          & 0.379          \\
CompletionFormer & 4.180          & 2.175          & 0.820          & 1.963          & 0.942          & 0.966          & 0.867          & 0.402          & 0.997          & 1.689          & 0.387          & 0.984          & 11.492         & 5.422          & 0.626          & 9.969          & 7.641          & 0.379          \\ \midrule
Ours+MiDaS       & 4.463          & 2.256          & 0.797          & 1.182          & 0.337          & 0.991          & 0.836          & 0.318          & 0.996          & 1.637          & 0.372          & 0.985          & 13.272         & 5.868          & 0.714          & 5.763          & 3.935          & 0.479          \\
Ours+KBR         & 3.780          & 1.719          & 0.874          & 1.180          & 0.339          & 0.993          & 0.944          & 0.257          & 0.995          & 1.689          & 0.381          & 0.985          & 12.032         & 5.479          & 0.705          & 5.790          & 4.291          & 0.483          \\
Ours             & \textbf{2.453} & \textbf{1.047} & \textbf{0.967} & \textbf{1.037} & \textbf{0.302} & \textbf{0.994} & \textbf{0.704} & \textbf{0.206} & \textbf{0.998} & \textbf{1.617} & \textbf{0.361} & \textbf{0.986} & \textbf{7.648} & \textbf{3.354} & \textbf{0.759} & \textbf{1.994} & \textbf{1.072} & \textbf{0.890} \\ \bottomrule
\end{tabular}%
}
\vspace{-3.8mm}
\caption{Case study of \textit{sparsity}, \textit{pattern}, and \textit{range} biases on the KITTI DC dataset.}
\label{tab:SPR_KITTI}
\end{table*}

\begin{figure*}[ht]
\centering
 \includegraphics[width=1.0\textwidth]{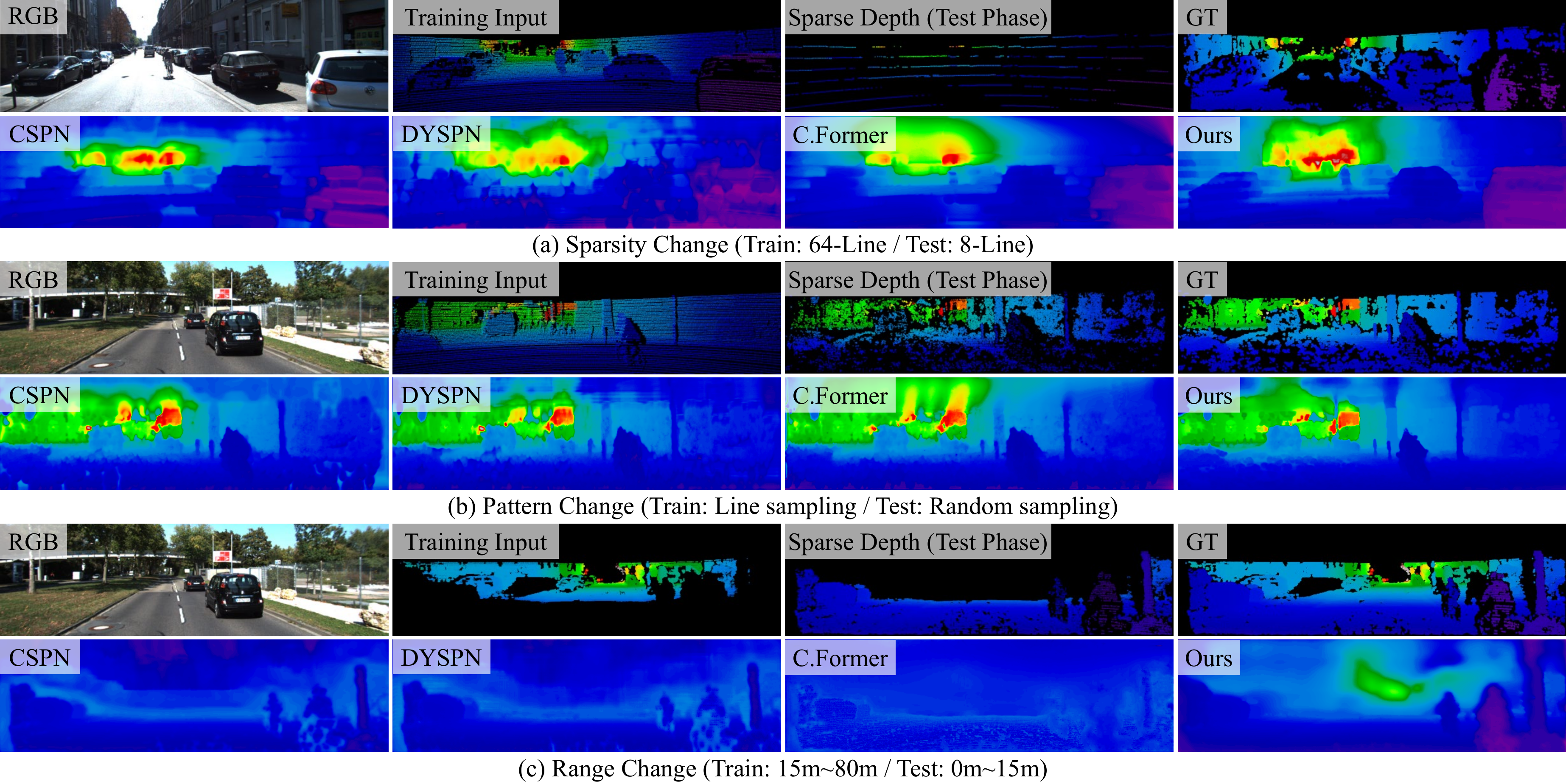}
 \vspace{-7mm}
 \caption{Qualitative results for the changes of measurement patterns, sparsity, and scan ranges.
 We visualize images, input examples in the training phase, sparse depths in the test phase, and GT in the first row.
}
 \label{fig:KITTI_SPR}
\end{figure*}



\subsection{Sparsity, Pattern and Range Biases}
\label{subsec:SPR}

To assess the effectiveness of our model against the sensor bias issues, we design experiments with varying conditions: sparsity (from 500 to 50 samples), patterns (from random to grid), and range changes (from 0m$\sim$3m to 3m$\sim$10m). We also design experiments for the outdoor scenario that vary across three key conditions: sparsity (from 64-Line to 8-Line), patterns (from line to random), and range changes (from 0m$\sim$15m to 15m$\sim$80m). 
For a fair comparison, all models do not conduct RDA for this experiment, including our proposed method.

\cref{tab:SPR_NYU,tab:SPR_KITTI} reveals that the previous methods face significant challenges on the bias issues. In addition, as demonstrated in~\cref{tab:kitti,tab:nyu}, the density changes also lead to significant performance drop, particularly the dense to sparse scenario. In contrast, to address the \textit{sparsity} bias, our prompt-based method constructs adaptive relations among pixels to properly propagate even in the changing conditions.



Next, we investigate the negative impact of \textit{pattern} bias. 
We observe that a model trained on depth data with a certain pattern suffers from limited generality due to the incompatibility with abundant representation learned for latent spaces of other depth patterns.
As shown in~\cref{tab:SPR_NYU}, we attribute this to positional information combined with image features. The comparison models, being continuously exposed to a fixed pattern like a grid shape, are limited to its generalization. When image and depth information are jointly represented, this issue is further exacerbated. 
On the other hand, from the results, we find out that random sampling offers benefits akin to augmentation effects, making the model generality better. Our method allows the transfer of depth information trained from various sparse patterns to the model, which provides the same effect as random sampling. 



Lastly, we check the \textit{range} bias. In the training phase, we only use depth data whose maximum depth range is 3m. All the models are tested using depth data whose min/max range of the depth distribution is set to [3m, 10m]. As shown in~\cref{tab:SPR_NYU}, it becomes evident that most methods exhibit poor generalization performance. Notably, the CompletionFormer~\cite{zhang2023completionformer} and NLSPN~\cite{nlspn} struggle to produce the general performance for the near and far regions. Our framework effectively tackles the challenge using the foundational model designed for monocular depth estimation. Based on the foundation model, which predicts relative depth maps for all pixels, our method infers absolute depth maps, which extends the sensor's limited scan ranges.

\cref{fig:KITTI_SPR} shows a significant distinction between our method and others in depth map reconstruction. While the comparison methods face challenges in accurately representing scene depth, especially in areas where input seeds are provided, our method excels in reconstructing the entire depth map. One notable observation is about the scenarios involving the changes in scanning ranges (\cref{fig:KITTI_SPR}-(c)). Our method uniquely overcomes the common bias problem. This better performance is attributed to the strengths of our foundation model's knowledge and the sensor-adaptive depth prompt.




\begin{table}[t]
\Huge
\centering
\renewcommand{\arraystretch}{0.9}%
\resizebox{1.0\columnwidth}{!}{
\begin{tabular}{c||c|c|c||c|c}
\toprule
                                       & {Sparsity}                       & {Pattern}                       & {Range}                         &  {Param.}                      & {Inference}                             \\ \cmidrule{2-6}
w/o SPN~Eq.(\textcolor{blue}{1})       & 0.498~/~0.334                    & 0.145~/~0.096                   & 0.546~/~0.379                   & 53.3M                             & 61.7ms                               \\
w/o Prompt~Eq.(\textcolor{blue}{2})    & 0.452~/~0.288                    & 0.301~/~0.207                   & 0.686~/~0.551                   & 49.7M                             & 54.9ms                               \\
w/o Pretrain~Eq.(\textcolor{blue}{3})  & 0.409~/~0.249                    & 0.118~/~0.049                   & 1.283~/~0.934                   & 326.9M                            & 64.4ms                               \\
~w/o LS-solver~Eq.(\textcolor{blue}{4})& 0.416~/~0.268                    & 0.118~/~0.052                   & 0.520~/~0.305                   & 53.4M                             & 61.3ms                               \\
w/ RDA                                 & \textbf{0.231}~/~\textbf{0.134}  & 0.113~/~0.046                   & 0.426~/~0.251                   & 53.4M                             & 63.9ms                               \\
Ours                                   & 0.400~/~0.242                    & \textbf{0.108}~/~\textbf{0.0420} & \textbf{0.206}~/~\textbf{0.108} & 53.4M                             & 63.9ms                              \\ \bottomrule
\end{tabular}
}
\vspace{-5mm}
\caption{Ablation study of our proposed methods (RMSE~/~MAE).}
\vspace{-3.5mm}
\label{tab:ablation_component}
\end{table}

\begin{table}[t]
\Huge
\centering
\resizebox{1.0\columnwidth}{!}{%
\begin{tabular}{c|c|c|c|c|c|c}
\toprule
                   & {NYU 100}                          & {NYU 8}                       & {NYU 1}                       & {KITTI 16}                        & {KITTI 4}                         & {KITTI 1}                         \\ \cmidrule{2-4} \cmidrule{5-7}
NLSPN              & \textbf{0.178}~/~0.089                    & 0.434~/~0.290          & 0.649~/~0.491          & 1.662~/~0.620          & 2.307~/~0.930          & 3.271~/~1.464                   \\ 
C.Former           & 0.182~/~0.090                    & 0.434~/~0.289          & 0.648~/~0.487          & 2.179~/~0.796          & 3.291~/~1.363          & 5.428~/~2.457                   \\ 
\textbf{Ours}      & \textbf{0.178}~/~\textbf{0.087}  & \textbf{0.367}~/~\textbf{0.246} & \textbf{0.404}~/~\textbf{0.285} & \textbf{1.351}~/~\textbf{0.413} & \textbf{1.951}~/~\textbf{0.763} & \textbf{2.823}~/~\textbf{1.268}          \\ \bottomrule
\end{tabular}%
}
\vspace{-8.5mm}
\caption{Adaption RDA method to other methods~(RMSE~/~MAE).}
\label{tab:RDA_abl}
\end{table}

\subsection{Ablation Study}
\label{subsec:Ablation}

\noindent\textbf{Adaption to Foundation Models.}\quad
To evaluate the versatility of our method with various foundational models, we replace our primary backbone with MiDaS~\cite{birkl2023midas} and KBR~\cite{spencer2023kick}. The MiDaS and KBR are developed for relative scale depth using large-scale datasets as well. \cref{tab:SPR_NYU} reveals that KBR outperforms other methods, including our own variant using the MiDaS backbone. We argue that the self-supervised training of KBR, unlike the supervised manner of MiDaS, provides a more generalizable feature space, dealing with challenging conditions~\cite{chen2020simple,wang2020understanding}.



\noindent\textbf{Component Ablation of the Proposed Method.}\quad
We perform an additional ablation study on each component of our model in~\cref{tab:ablation_component}. The study reveals that the RDA method notably reduces the sparsity bias~(w/ RDA). For the range bias, the pre-trained knowledge from the backbone contributes to performance improvement~(w/o Pretrain). Our depth-oriented prompt engineering contributes to the overall performance~(w/o Prompt). Additionally, the results of LS solver Eq.(4) show that the sensor biases are not addressed via the naive scale fitting, but are solved by the combinations of our components. As a solution, our idea is to exploit the SPN module to use the initial dense depth, which is aligned relative depth from the backbone with sparse absolute-scale depth~(w/o SPN). 


\noindent\textbf{Random Depth Augmentation.}\quad
RDA is an effective strategy to mitigate the issue of sparsity bias. To evaluate its compatibility and adaptability with other methods, we conduct an ablation study incorporating RDA into NLSPN~\cite{nlspn}, and CompletionFormer~\cite{zhang2023completionformer}. The results, as described in~\cref{tab:RDA_abl}, demonstrate notable performance improvements in sparse input scenarios. This highlights that the RDA not only naturally improves the models' ability to generalize across different levels of data sparsity, but also becomes more effective when used together with prompt engineering. 

\subsection{Zero-shot Inference on Commercial Sensors}
\vspace{-2mm}

We verify our method's zero-shot generality by testing it on different datasets without any additional training. We use our model trained on NYUv2~\cite{NYUv2} and KITTI DC~\cite{kitti_DC_benchmark} datasets, then apply it to dataset taken from various sensors such as Apple LiDAR~\cite{luetzenburg2021evaluation}, Intel RealSense~\cite{keselman2017intel}, and 32-Line Velodyne LiDAR~\cite{schwarz2010mapping}. Here, for the Apple LiDAR dataset, we directly capture a set of indoor images using iPad Pro. As shown in~\cref{fig:teaser}, our method is applicable for Apple LiDAR compared to ARKit~\cite{luetzenburg2021evaluation}, which is a built-in framework   on iOS devices. Additionally, it shows better generality with consistent results in the VOID dataset~\cite{wong2020unsupervised}, collected using a stereo sensor in RealSense. Remarkably, our model, initially trained on 64-Line Velodyne LiDAR, excels in handling the NuScenes dataset~\cite{wong2020unsupervised} captured with fewer channel LiDAR over the second best approach in \cref{subsec :Quantitative results}. Note that we describe details about the experimental setup,
more quantitative and qualitative results, and further analysis in the supplemental materials.

\label{subsec: Zero-shot Inference on Commercial Sensors}
\section{Conclusion}
We introduce a novel depth prompting technique, leveraging large-scale pre-trained models for high-fidelity depth estimation in metric scale. This approach significantly addresses the challenges of well-known sensor biases associated with fixed sensor densities, patterns, and limitation of range and enables sensor-agnostic depth prediction. Through the comprehensive experiments, we demonstrate the stability and generality of our proposed method, showcasing its superiority over existing methodologies. Furthermore, we verify our method on a variety of real-world scenarios through zero/few-shot inference across diverse sensor types.



%

\vspace{3mm}
\fontsize{8.5}{9.5}\selectfont{\noindent\textbf{Acknowledgement} This research was supported by 'Project for Science and Technology Opens the Future of the Region' program through the INNOPOLIS FOUNDATION funded by Ministry of Science and ICT (Project Number: 2022-DD-UP-0312), GIST-MIT Research Collaboration grant funded by the GIST in 2024, the Ministry of Trade, Industry and Energy (MOTIE) and Korea Institute for Advancement of Technology (KIAT) through the International Cooperative R$\&$D program in part (P0019797) and the Korea Agency for Infrastructure Technology Advancement(KAIA) grant funded by the Ministry of Land Infrastructure and Transport (Grant RS-2023-00256888).}

\clearpage
{
    \small
    \bibliographystyle{ieeenat_fullname}
    \bibliography{main}
}


\end{document}